\documentclass[graybox]{svmult}

\usepackage[table]{xcolor}
\usepackage{mathptmx}       % selects Times Roman as basic font
\usepackage{helvet}         % selects Helvetica as sans-serif font
\usepackage{courier}        % selects Courier as typewriter font
\usepackage{type1cm}        % activate if the above 3 fonts are
\usepackage{makeidx}         % allows index generation
\usepackage{graphicx}        % standard LaTeX graphics tool
\usepackage{multicol}        % used for the two-column index
\usepackage[bottom]{footmisc}% places footnotes at page bottom
\usepackage{subfigure}
\usepackage{multirow}
\usepackage{amsmath,amssymb}

\makeindex             % used for the subject index
\begin{document}

\title*{A Multi-View Ensemble Classification Model for Clinically Actionable Genetic Mutations}
% Use \titlerunning{Short Title} for an abbreviated version of
% your contribution title if the original one is too long
\author{Xi Sheryl Zhang$^{a}$, Dandi Chen$^{a}$, Yongjun Zhu$^{a}$, Chao Che$^{a,b}$, Chang Su$^{c}$, Sendong Zhao$^{d}$, Xu Min$^{a,e}$, Fei Wang$^{a,*}$\\
}
\authorrunning{Zhang et al.}
\titlerunning{A Multi-View Ensemble Classification Model for Genetic Mutations}
% Use \authorrunning{Short Title} for an abbreviated version of
% your contribution title if the original one is too long
\institute{$^{a}$Department of Healthcare Policy and Research, Weill Cornell Medicine, Cornell University. $^{b}$Key Laboratory of Advanced Design and Intelligent Computing, Ministry of Education, Dalian University. $^{c}$School of Electronic and Information Engineering, Xi’an Jiaotong University, China. $^{d}$Research Center for Social Computing and Information Retrieval, Harbin Institute of Technology, China. $^{e}$Department of Computer Science and Engineering, Tsinghua University, China.\\
$^*$Corresponding Author. Email: few2001@med.cornell.edu}
%
% Use the package "url.sty" to avoid
% problems with special characters
% used in your e-mail or web address
%
\maketitle

\abstract{This paper presents details of our winning solutions to the task IV of NIPS 2017 Competition Track entitled \textit{Classifying Clinically Actionable Genetic Mutations}. The machine learning task aims to classify genetic mutations based on text evidence from clinical literature with promising performance. We develop a novel multi-view machine learning framework with ensemble classification models to solve the problem. During the Challenge, feature combinations derived from three views including document view, entity text view, and entity name view, which complements each other, are comprehensively explored. As the final solution, we submitted an ensemble of nine basic gradient boosting models which shows the best performance in the evaluation. The approach scores $0.5506$ and $0.6694$ in terms of logarithmic loss on a fixed split in stage-$1$ testing phase and $5$-fold cross validation respectively, which also makes us ranked as a top-$1$ team\footnote{https://www.mskcc.org/trending-topics/msk-advances-its-ai-machine-learning-nips-2017} out of more than $1300$ solutions in NIPS $2017$ Competition Track IV.}

\section{Introduction}
\label{sec:1}

The NIPS $2017$ Competition Track IV arises from gene mutation classification \footnote{https://nips.cc/Conferences/2017/Schedule?showEvent=8748} using Natural Language Processing (NLP). Gene mutation classification, which is one of the important problems in Precision Medicine, aims at distinguishing the mutations that contribute to tumor growth (drivers) from the neutral mutations (passengers). Identifying types of gene mutations is helpful in determining emerging molecular tumors and finding drugs that can treat them~\cite{Tutorials}. In order to classify clinically actionable genetic mutations, the related biomedical scholarly articles are a trustworthy knowledge source. Currently, this interpretation of genetic mutations is being done manually. This is a quite time-consuming task where a clinical pathologist has to manually review and classify every single genetic mutation based on evidence from articles\footnote{https://www.kaggle.com/c/msk-redefining-cancer-treatment}.

The Competition Track releases a dataset of oncogenes~\cite{he2005microrna}, along with their corresponding mutations and related articles obtained from PubMed, an online biomedical literature repository. The goal is to design machine learning approaches which can predict class labels for gene mutation samples with acceptable accuracy. The target classes have been predefined by the Challenge organizer Memorial Sloan Kettering Cancer Center (MSKCC). Specifically, they are ``Gain-of-function'', ``Likely Gain-of-function'', ``Loss-of-function'', ``Likely Loss-of-function'', ``Neutral'', ``Likely Neutral'', ``Switch-of-function'', ``Likely Switch-of-function'', and ``Inconclusive''. Therefore, it is a multi-class classification task.  

\begin{figure}
\centering
\includegraphics[width=1\textwidth]{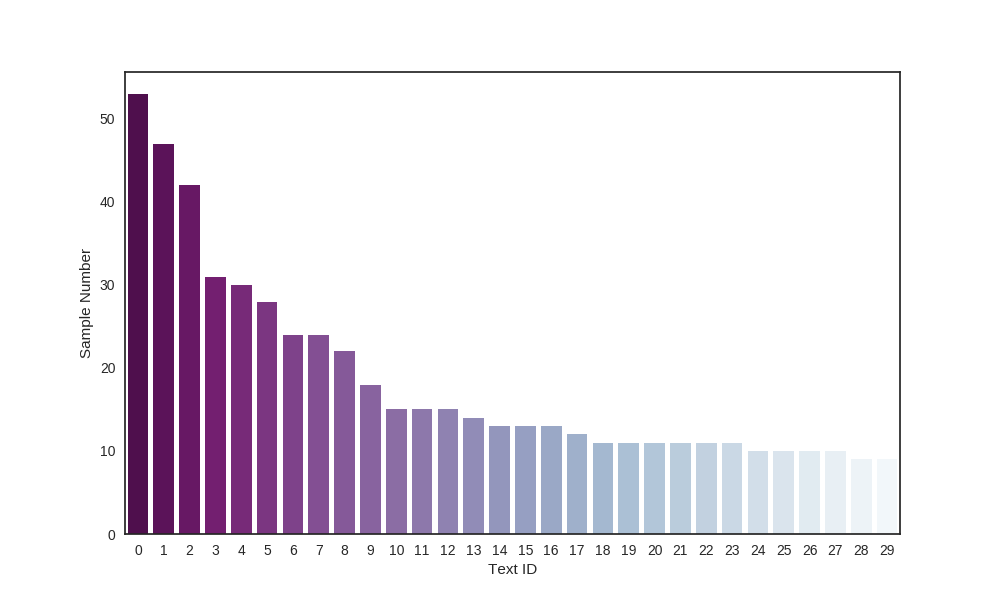}
\caption{\label{fig:problem1} Distribution of the counts of common text that are shared by different samples. The head of the distribution is shown here. When we give all the observed text a unique id, the most common text is used more than 50 times by gene mutation samples.}
\end{figure}

\begin{figure}
\centering
\includegraphics[width=1\textwidth]{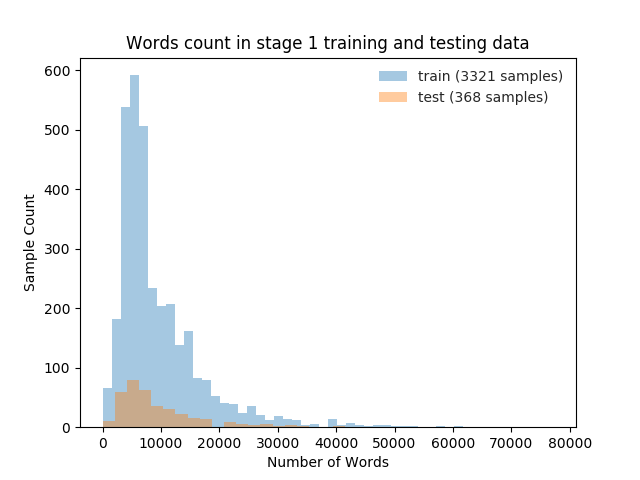}
\caption{\label{fig:problem2} Distribution of the text entry lengths. The median value of word count per text is 6,743 while the maximum word count in a text is 77,202.}
\end{figure}

Basically, the competition can be viewed as a text classification task based on clinical descriptions of gene mutations. However, the problem is more challenging than traditional document classification problem that is handled with NLP benchmarks in many aspects. Our useful observations about the difficulties are summarized as follows:

\begin{itemize}
\item Different samples may share the same text entry, while their class labels are entirely different. From the statistics shown in Fig.~\ref{fig:problem1}, there are a bunch of cases in which various samples share the same text. Instead of mining knowledge from original documents, more evidence from other perspective is necessary. 

\item Each sample is associated with an extremely long document with a large amount of noisy information that makes the problem difficult. As the word count distribution shown in Fig.~\ref{fig:problem2}, documents generally contain much more sentences than the normal text classification datasets~\cite{textclassification}.  

\item While a gene and its corresponding label distribution over classes could be a great hint in the prediction, the fact that there are only a few overlapped samples in training and testing set makes the distributional information useless. Basically, we can only summarize effective features from characters through entity names. 
\end{itemize} 
   
In order to deal with above challenges, a multi-view ensemble classification framework is proposed. Concretely, we extract prior knowledge about genes and mutations globally from the sentences mentioning the specific genes or mutations in the text collection. Hence, we are able to design text features not only for the original document view but also for the entity (gene/mutation) text view to address the first two difficulties above-mentioned. To make full use of the entity name information, the third view for names is also explored using word embedding or character-level n-gram encoding. Furthermore, we combine features derived from three views to implement basic classification models and exploit features from each view complements to each other. After that, we ensemble the basic models together by several strategies to boost the final accuracy. The data and source code are available at \url{https://github.com/sheryl-ai/NeurIPS17}.   

The rest of the paper is organized as follows. Section 2 introduces main notations, the validation dataset, and the evaluation metric. In Section 3, the multi-view text mining approach is explained. Model ensemble methods are presented in Section 4. Empirical study and analysis are provided in Section 5. Eventually, several conclusions are given in Section 6.

\section{Preliminary}
\label{sec:2}

\subsection{Notations}
\begin{table}[t]
\tabcolsep 0.36in
\renewcommand\arraystretch{1.3}
\caption{main notations used in this paper}
\begin{tabular}{ll}
\hline
\textbf{Symbols}    \qquad\qquad     &\textbf{Definition}  \\
\hline
$\mathbf{x}$                   \qquad\qquad        &feature vector     \\
$\mathbf{x}^{D}$               \qquad\qquad        &feature vector in original document view    \\
$\mathbf{x}^{ET}$              \qquad\qquad        &feature vector in entity text view    \\
$\mathbf{x}^{ET}_{g}$          \qquad\qquad        &gene feature vector in entity text view     \\
$\mathbf{x}^{ET}_{m}$           \qquad\qquad       &mutation feature vector in entity text view     \\
$\mathbf{x}^{EN}$              \qquad\qquad        &feature vector in entity name view    \\
$\mathbf{x}^{EN}_{g}$          \qquad\qquad        &gene feature vector in entity name view     \\
$\mathbf{x}^{EN}_{m}$           \qquad\qquad       &mutation feature vector in entity name view     \\
$N$             \qquad\qquad        &the number of samples    \\
$M$             \qquad\qquad        &the number of classes    \\
$y_{ij}$        \qquad\qquad        &binary indicator whether label $j$ is true for sample $i$ \\
$p_{ij}$        \qquad\qquad        &predicted probability of assigning label $j$ to sample $i$ \\
$\mathcal{T}^{r}$   \qquad\qquad     &training set   \\
$\mathcal{T}^{v}$   \qquad\qquad     &validation set \\
$\mathcal{T}^{s}$   \qquad\qquad     &testing set \\
$\hat{p}^{v}_{ij}$  \qquad\qquad     &predicted probability for validation set data \\
$\hat{p}^{s}_{ij}$  \qquad\qquad     &predicted probability for testing set data \\
$\alpha$        \qquad\qquad         &linear combination parameter for basic models\\
\hline
\end{tabular}
\label{table:notations}
\end{table}

Table~\ref{table:notations} lists some main notations used throughout the paper. In the paper, genes, mutations, and their corresponding documents are respectively denoted by $g$, $m$, and $d$. Each sample is constructed by a triplet $<gene, mutation, document>$. The feature vector generated for each sample is denoted as the vector $\mathbf{x}$. Feature vectors in the three views are represented as $\mathbf{x}^{D}$, $\mathbf{x}^{ET}$, and $\mathbf{x}^{EN}$ respectively. With notations presented in Table~\ref{table:notations}, our problem can be explicitly defined as:
\begin{description}[definition]
\item[Definition]{Given sample sets $\{<g_{i}, m_{i}, d_{i}>\}_{i=1}^{N}$, our aim is to generate feature vectors $\{\mathbf{x}_{i}^{D}, \mathbf{x}_{i}^{ET}, \mathbf{x}_{i}^{EN}\}_{i=1}^{N}$ in multiple views, so that probabilities of label assignment over $M$ possible class can be predicted.}
\end{description}

\subsection{Validation Set}

\begin{table}[t]
\tabcolsep 0.4in
\renewcommand\arraystretch{1.3}
\caption{Statistics of the stage-$1$ datasets}
\begin{tabular}{lll}
\hline
  \qquad\qquad  &\textbf{training}  &\textbf{validation} \\
\hline
\# of samples            \qquad\qquad   &3,321 &368   \\
\# of unique genes       \qquad\qquad   &264   &140  \\
\# of unique mutations   \qquad\qquad   &2,996 &328   \\
\hline
\end{tabular}
\label{table:datastats}
\end{table}

The Challenge consists of two stages and released a training set and a validation set in stage-$1$. During stage-$1$, the labels of the validation set are unknown and participants can verify the performances of their models online by submitting the classification results of the validation set. The stage-$1$ of this Challenge lasted for a couple of weeks, while the ongoing stage-$2$ was held in the final week. In stage-$2$, the stage-$1$ training data, validation data, and new online test data without labels are given. The stage-$1$ training set contains $3,321$ gene mutation samples with $264$ unique genes and $2,996$ unique mutations. The validation set contains $368$ gene mutation samples with $140$ unique genes and $328$ unique mutations. In total, we have $3,689$ training samples including $269$ unique genes and $3,309$ unique mutations. The detailed data statistics for the training set and the validation set can be found in Table~\ref{table:datastats}.

The stage-$1$ validation data is used to generate the rankings on the Leaderboard of the first stage. On the one hand, we perform offline validation without submitting classification results using the validation set. On the other hand, it can be used to extend the size of training set during the second stage of this competition. In this work, we denote the stage-$1$ training set ($3,321$ samples) by $\mathcal{T}^{r}$, and denote the stage-$1$ validation set ($368$ samples) by $\mathcal{T}^{v}$. The online testing set of stage-$2$ for submission is denoted by $\mathcal{T}^{s}$.  

\subsection{Evaluation Metric}
The Challenge adopts logarithmic loss as the evaluation metric. It measures performance of a multi-class classification model where the prediction is a probability distribution over classes between $0$ and $1$. Mathematically, the Logloss is defined as:

\begin{equation}
-\frac{1}{N}\Sigma_{i=1}^{N}\Sigma_{j=1}^{M} y_{ij}log(p_{ij}) 
\label{equation:metric}
\end{equation}
where $N$ and $M$ are the number of samples and the number of the possible class label, respectively. $y_{ij}$ is a binary indicator of whether or not label $j$ is the correct classification for sample $i$, and $p_{ij}$ is the output probability of assigning label $j$ to instance $i$. By minimizing log loss, the accuracy of the classifier is maximized. In other words, a smaller value of the metric indicates a more accurate classifier.

\section{The Proposed Approach}
\label{sec:3}
% \subsection{Framework}
Given the gene mutations and their associated articles, a straightforward approach is to extract features directly from the documents and entity names. As we introduced, this approach might suffer the fact that two samples share same text but have different class labels. Considering a gene \textit{BRCA1}, it owns two possible mutations: \textit{T1773I} and \textit{M1663L} in two different samples, with their gene mutation types are ``Likely Loss-of-function'' and ``Likely Neutral'', respectively. The article descriptions, however, are exactly same for the two samples. The straightforward document classification approach cannot work well in this case, since it is fairly difficult for the classifier to categorize the samples into correct classes only via the name of mutations (normally a few characters construct the names).    

\begin{figure}[t]
\centering
\includegraphics[width=1\textwidth]{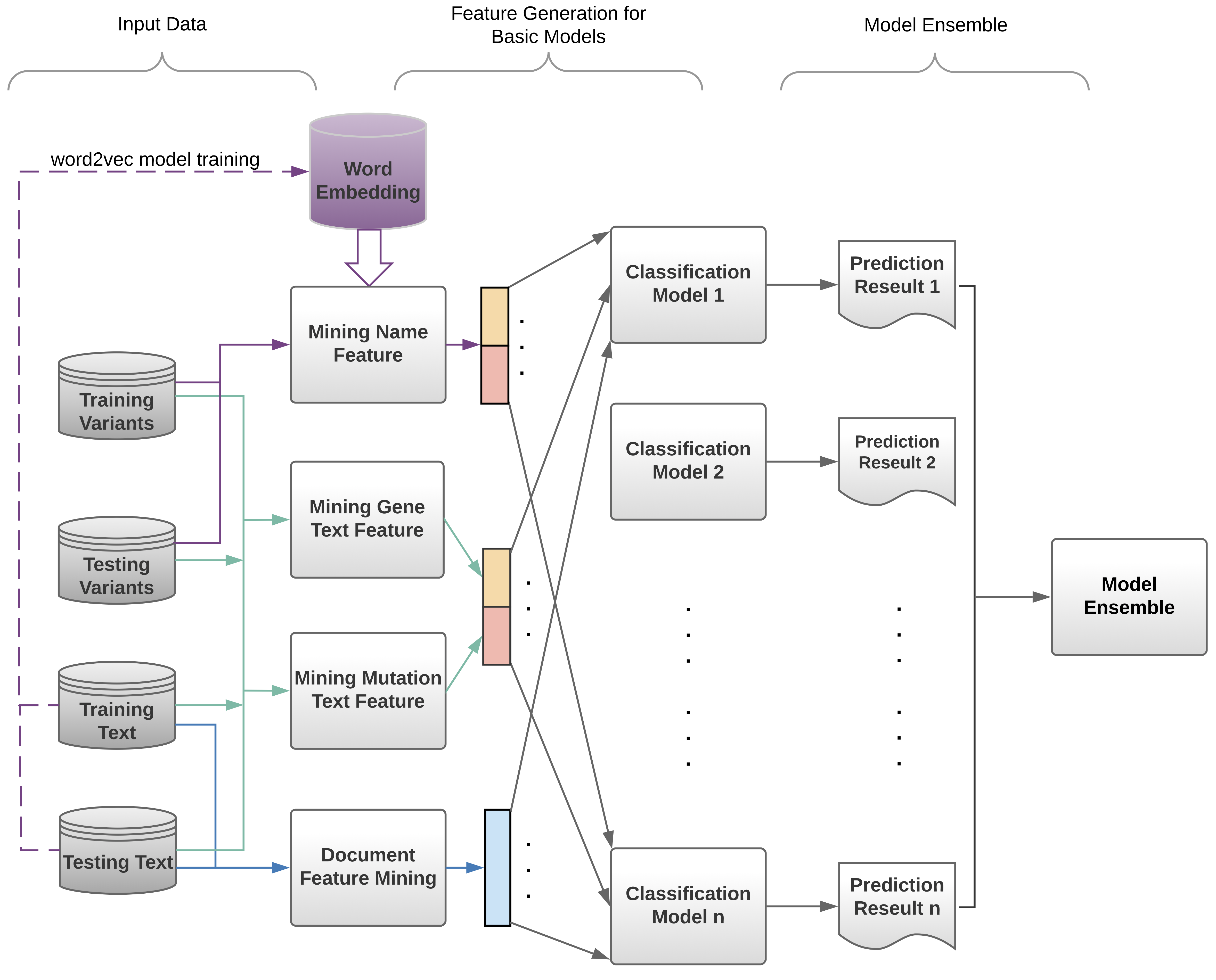}
\caption{\label{fig:framework} The classification framework (best viewed in color). Four data files are released by the Challenge: training/testing variants and training/testing text. The three colored arrows from data files to feature mining modules indicate three aspects of feature engineering. Document features are only derived from text data; entity text features need both variants and text; entity name features derive from variants as well as text data (Word embedding model is also trained using the given text).}
\label{framework}
\end{figure}

Fig.~\ref{fig:framework} presents an overview of our multi-view framework for solving this problem. The original input data includes training and testing variants (the name information of gene mutations), training and testing texts (the associated articles\footnote{We use articles and documents interchangably in this paper}). In our solution, we perform feature extraction and engineering from the following three views:
\begin{itemize}
\item Document view: original documents associated with gene mutation samples (denoted by blue arrows in Fig.~\ref{fig:framework});
\item Entity text view: sentences globally extracted from the document collection associated with genes and mutations (denoted by green arrows in Fig.~\ref{fig:framework});
\item Entity name view: characters of the gene names and mutation names (denoted by purple arrows in Fig.~\ref{fig:framework}).
\end{itemize} 

After feature engineering, we first concatenate the gene text feature with mutation text feature to represent each sample. In particular, $\mathbf{x}_{g}^{ET}$ and $\mathbf{x}_{m}^{ET}$ are concatenated to form the feature vector of entity text $\mathbf{x}^{ET}=\mathbf{x}_{g}^{ET}\|\mathbf{x}_{m}^{ET}$, where $\|$ denotes concatenation operation. Similarly, the feature vector of entity name is formed by concatenation $\mathbf{x}^{EN}=\mathbf{x}_{g}^{EN}\|\mathbf{x}_{m}^{EN}$. Then features from three views are combined to train multiple classification models and generate multiple prediction results. Various combination schemes are explored to decide the feature sets with the best accuracy (see Section~\ref{sec:4}). The feature generation and combination will be introduced in the following sections. 

\subsection{Document View}

\subsubsection{Domain Knowledge} 
Domain knowledge usually provides extra pieces of information for classification task. To incorporate biomedical domain knowledge, feature dimensions including bioentities and keywords are extracted. 

Genes and mutations may have alias in PubMed articles. Also, quite a lot bioentities appear in the text but not be included in samples. A proper utilization of the bioentity information is an important part of a successful solution. Thanks to a Named Entity Recognition (NER) tool PubTator~\cite{wei2013pubtator}, we can extract the entity dictionary for the entities in the text data. The PubTator is used to enrich the dictionaries of genes and mutations using the abstracts of the related PubMed articles. The tool includes GeneTUKit~\cite{huang2011genetukit}, GenNorm~\cite{gennorm} and tmVar~\cite{wei2013tmvar}. Finally, we obtain bioentities containing $2,668$ chemicals, $2,486$ diseases, $6,987$ genes, and $2,486$ mutations.  

In addition to the document corpus provided by this Challenge, we also built a dictionary by \emph{Keywords} extracted from related PubMed articles obtained from OncoKB\footnote{http://oncokb.org/}. The underlying assumption is that the keywords detected from titles of the related articles are the essential terms in the research domain. In particular, the keywords are extracted from the titles of those articles by removing the stop words and punctuations. The keywords dictionary has $3,379$ unique words. 

\subsubsection{Document-Level Feature} 
While traditional feature engineering will always be staples of machine learning pipelines, representation learning has emerged as an alternative approach to feature extraction. In order to represent a document by natural language modeling, paragraph vectors or Doc2Vec~\cite{doc2vec} is exploited. Doc2Vec can be viewed as a generalization of Word2Vec~\cite{Word2Vec} approach. In addition to considering context words, it considers the specific document when predicting a target word, and thus it can exploit the word ordering along with their semantic relationships. With a trained Doc2Vec model, we can get a vector representation with a fixed length for any given document with arbitrary lengths. 

Doc2Vec provides two training strategies to model the context in documents, which are distributed memory model (PV-DM) and distributed bag-of-word model (PV-DBOW). PV denotes paragraph vector here. Given sequences of training words in paragraphs, the PV-DM model is trained to get paragraph vectors, word vectors, softmax weights and bias to maximize the probability of seen texts. The difference between the two versions are: PV-DM simultaneously uses the context words and a paragraph matrix to predict the next word while PV-DBOW ignores the context words in the input but uses the parameter matrix to predict words randomly sampled from the paragraphs, which leads to less storage. As recommended in~\cite{doc2vec}, we combine the outputs of PV-DM with PV-DBOW together to achieve the best performance (concatenation of $150$ dimensions for PV-DM and $250$ dimensions for PV-DBOW).             

\subsubsection{Sentence-Level Feature} 
When it comes to extracting features from very noisy long documents, filtering sentences might be a choice to obtain effective knowledge. Regarding sentences mentioning the genes or mutations as key sentences, the context of key sentences is also used to capture the useful information. The basic assumption behind is that words in the same context tend to have similar meanings. For the reason that the articles have sufficient sentences to satisfy the distributional hypothesis theory~\cite{harris}, we extend the concept "word" to "sentence" to form contexts. Considering a key sentence $s_{t}$ and its Term Frequency-Inverse Document Frequency (TF-IDF) feature vector $\mathbf{x}_{s,t}$, the context can be represented as concatenation: $\mathbf{x}_{s,t-1}\|\mathbf{x}_{s,t}\|\mathbf{x}_{s,t+1}$ when the window size is set as $3$. Then the representation for documents in samples can be calculated by averaging the key contexts. Here we adopt average values and call the defined feature as sentence-level TF-IDF.     

\subsubsection{Word-Level Feature} 
Nouns, verbs, adjectives, and adverbs are four major word types we considered in Part-of-Speech (PoS) Tagging~\cite{postagging, wordnet}. In the scenario of genetic mutation classification, nouns could be the names of \emph{proteins}, \emph{diseases}, \emph{drugs}, and so on, which serve as important clues for predicting mutation classes. The verb type includes most of the words referring to actions and processes such as \emph{interact}, \emph{affect}, and \emph{detect}. In addition, adjectives are considered since they reflect properties of nouns while adverbs might semantically reflect some discoveries or conclusions like \emph{dramatically}, or \emph{consistently}. Our method takes all of the word tokens as input with preprocessing steps including filtering out stop words and punctuations, stemming and lemmatization, and PoS tagging. Then a dictionary with $9,868$ words of all the four types is constructed.   
 
TF-IDF is one of the common used measures that computes the importance of each word in a document~\cite{tfidf}. Given a collection of $N$ documents. TF-IDF assigns to term $t$ a weight in document $d$ given by $\mathrm{tfidf}_{td}=\mathrm{tf}_{td}\times \mathrm{idf}_{t}$, where inverse document frequency can be defined as $\mathrm{idf}_t=log\frac{N}{\mathrm{df}_t}$. Document frequency $\mathrm{df}_t$ means the number of documents in the collection that contain term $t$. Our new strategy is to embed the discriminative power of each term. Intuitively the $\mathrm{idf}_{t}$ should be calculated by class frequency, that is, $\mathrm{idf}_t=log\frac{M}{\mathrm{cf}_t}$, where $M$ is the number of class and $\mathrm{cf}_t$ the number of classes that contain a term $t$.   

In addition to the designed novel TF-IDF, we compare several different value computation methods such as word counts, TF, and TF-IDF based on bag-of-words for a better performance.    

\begin{table}[t]
\tabcolsep 0.62in
\renewcommand\arraystretch{1.3}
\caption{Dimensions of sentence-level and word-level features before and after using dimension reduction (the statistics is computed in document view).}
\begin{tabular}{lc}
\hline
\textbf{Features}    \qquad\qquad     &\textbf{Dimension}  \\
\hline
n/v./adj./adv. counts       \qquad\qquad        &9,868     \\
ngram                       \qquad\qquad        &9,473,363    \\
sentence-level TFIDF        \qquad\qquad        &28,368    \\
term frequency              \qquad\qquad        &9,456     \\
% bioentity counts            \qquad\qquad            &10,022    \\
% keyword counts              \qquad\qquad            &3,379    \\
\hline
n/v./adj./adv. counts+NMF   \qquad\qquad            &60     \\
ngram+NMF                   \qquad\qquad            &120 \\
sentence-level TFIDF+SVD    \qquad\qquad            &100    \\
term frequency+LDA          \qquad\qquad            &50     \\
\hline
\end{tabular}
\label{table:featuredim}
\end{table}

\subsubsection{Dimension Reduction} 
In general, original features based on bag-of-words or bag-of-n-grams may have thousands of dimensions. For example, the dimension can reach more than $9$ million when we adopt unigram/bigram/trigram simultaneously. The designed features and their corresponding dimensions are shown in the Table~\ref{table:featuredim}. To solve the problem of high-dimensional input, dimension reduction for feature vectors is taken into account. Dimension reduction is the process of reducing the number of features~\cite{DimReduction}. On the one hand, it can help the classification models improve their computational efficiency. On the other hand, it can reduce the noise and sparsity of the raw features. Popular dimension reduction techniques including Singular-Value Decomposition (SVD)~\cite{SVD}, Non-negative Matrix Factorization (NMF)~\cite{NMF}, and Latent Dirichlet Allocation (LDA)~\cite{LDA} have demonstrated promising results on multiple text analysis tasks. Hence, SVD, NMF, and LDA are implemented in our solution. 

We combine the bag-of-words or bag-of-n-grams with SVD, NMF, or LDA, and choose the feature combinations according to their achieved performance. Finally, we obtain feature vectors with dimensionality of $50$, $60$, $100$, and $120$. Table~\ref{table:SingleFeature} reports the detailed settings for dimension reduction. The feature vector from document view is represented as $\mathbf{x}^{D}$.

\begin{figure}[t]
\subfigure[]{
\begin{minipage}[b]{0.5\linewidth}
\centering
\includegraphics[width=\textwidth]{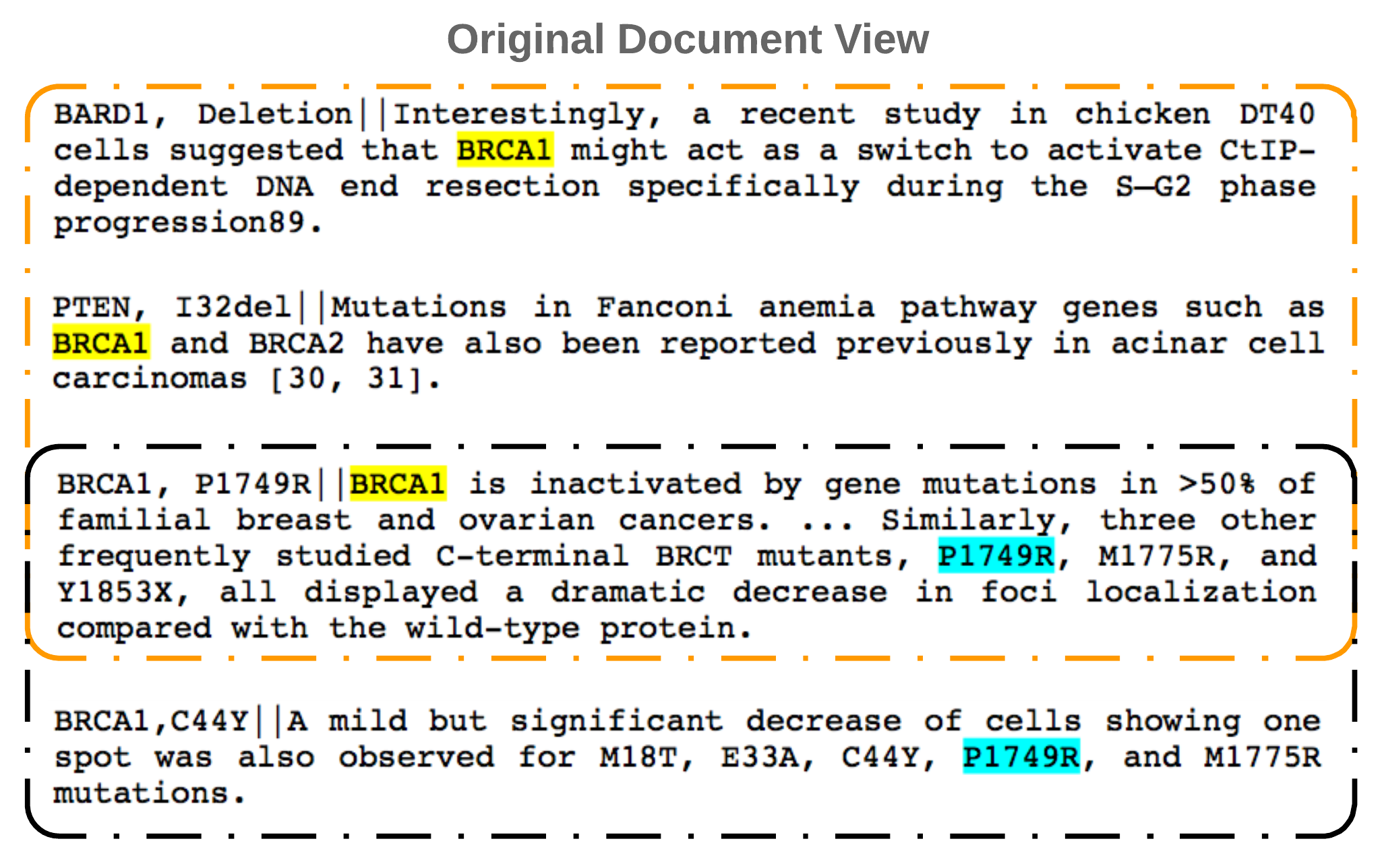}
\end{minipage}}%
\subfigure[]{
\begin{minipage}[b]{0.5\linewidth}
\centering
\includegraphics[width=\textwidth]{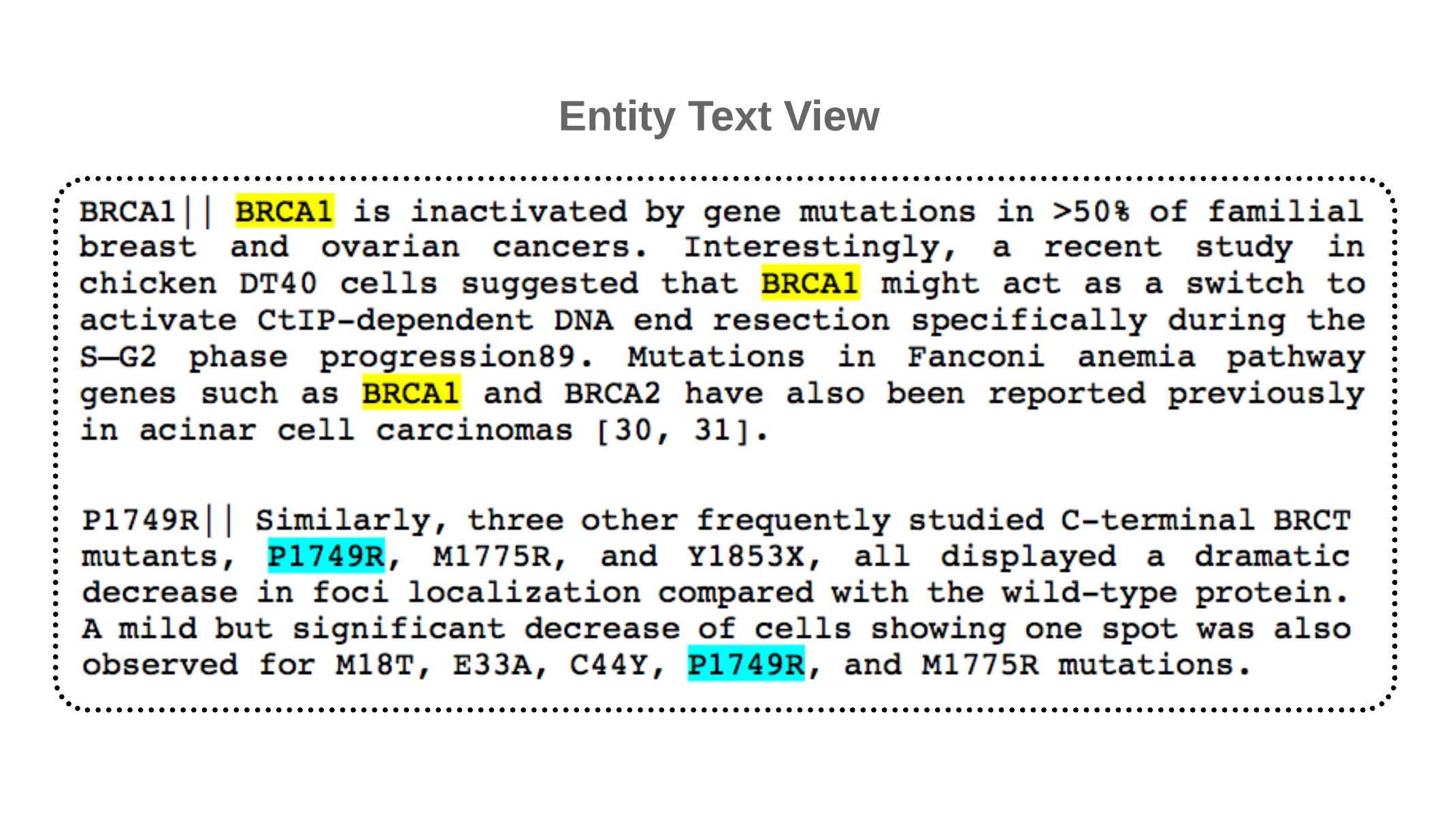}
\end{minipage}}
\vspace{-0.5cm}
\caption{\label{figure:ViewConstruction} A toy example of constructing the entity text view. Original document view is the data provided by the Challenge. Entity text view is the extracted sentences from the overall documents globally that mentioned the specific gene or mutation. The entity texts for gene mutations are collected separately. The given example illustrates the view construction of a gene \textit{BRCA1} and its mutation \textit{P1749R}. Then we can understand the knowledge not only from the document view but also from the entity text view.}
\label{figure:ViewConstruction}
\centering
\end{figure}

\subsection{Entity Text View}
As we mentioned before, documents are too long and it would be helpful to analyze the view of texts containing individual genes or mutations. Correspondingly, we developed a two-step method including view construction and feature generation in the procedure of entity text view. 

\subsubsection{View Construction}
Fig.~\ref{figure:ViewConstruction} shows an illustrative example of entity text extraction. Basically, we firstly match strings with the names of gene or mutation in documents and then extract the sentences containing those strings. A trie tree-like fast string search algorithm named Aho-Corasick (AC) automaton~\cite{AC} is adopted. The complexity of the AC algorithm is linear $\mathcal{O}(n + m + z)$, where $n$, $m$ and $z$ are the length of the strings, the length of the texts, and the number of output matches, respectively. Without AC automaton, the time complexity of exact string matching is $\mathcal{O}(n + km)$ where $k$ is the number of patterns (genes or mutations in our scenario) that need to be found. Hence, it could take days to extract sentences with thousands of genes or mutations from original text to entity text, which is computationally prohibitive. As the computational complexity shown, AC automation is capable of solving the efficiency problem to a large extent.

\subsubsection{Feature Generation} 
Once the sentences containing the names of gene or mutation are extracted, we collect all sentences mentioning a specific gene or a specific mutation as a separate entity text. Then the document feature engineering approaches introduced in the last subsection can be applied to these entity texts to generate feature vectors. Fortunately, both sentence-level features and word-level features show impressive performance on the top of entity texts. Note that the sentence-level TF-IDF is changed using the key sentences instead of context. Nevertheless, the document-level assumption of paragraph vector is not consistent with the entity view, since it lacks rationale to optimize the paragraph vector on the text without orderly sentences. 

We concatenate the gene feature vector $\mathbf{x}^{ET}_{g}$ and mutation feature vector $\mathbf{x}^{ET}_{m}$ to get the combined feature vector $\mathbf{x}^{ET}=\mathbf{x}^{ET}_{g}||\mathbf{x}^{ET}_{m}$ for a specific gene mutation sample, as shown in Fig.~\ref{fig:framework}. For instance, suppose a gene and a mutation are given, the n-gram feature for the given sample with a gene and a mutation is generated separately, on the basis of their corresponding extracted text. Then the concatenated n-gram vector can be used to represent the sample. The feature vector generated from entity text view is represented as $\mathbf{x}^{ET}$.        
 
\subsection{Entity Name View}
Though most of the gene names and mutation names are short and only consist of few characters and numbers, the name itself contains useful information for classification. Two encoding approaches are designed to capture patterns from names, which are character-level n-gram and word embedding. 

\subsubsection{Character-Level Feature}
Unlike word-level n-gram, we can set a large $n$ ($n=8$) as names are typically short strings. As a consequence, the feature dimension is extremely high. We adopt SVD to reduce the dimensionality to $20$. The other encoding approach uses label encoder to transform the letters and numbers in gene or mutation name into digital labels ($112$ unique labels in total) that can be used as feature directly.

\subsubsection{Word Embedding Feature}
Word embedding is a technique aiming at representing (embedding) words in a continuous vector space where semantically similar words are mapped to nearby points. Representative word embedding techniques include Word2Vec~\cite{Word2Vec} and GloVe~\cite{GloVe}. The trained word embedding models can offer us feature vector representations for each specific gene or mutation according to their names. In this task, we choose Word2Vec (Skip-Gram)~\cite{skip-gram} because both Word2Vec and GloVe achieve similar classification performance during the evaluation. The feature dimension of gene or mutation name vectors is set as $200$ according to cross-validation.  

Similar to entity text view, the feature vector extracted from entity name view is concatenated by gene feature vector $\mathbf{x}^{EN}_{g}$ and mutation feature vector $\mathbf{x}^{EN}_{m}$, that is $\mathbf{x}^{EN}=\mathbf{x}^{EN}_{g}||\mathbf{x}^{EN}_{m}$. The feature vector generated from entity name view is represented as $\mathbf{x}^{EN}$.   

\subsection{Classifiers}
Gradient Boosting Decision Tree (GBDT) ~\cite{friedman2001greedy} is a famous machine learning technique for regression and classification problems. Based on boosting, it aims to find an optimal model $f(\mathbf{x})$ that satisfies the following equation: 
\begin{equation}
 \hat{f}(\mathbf{x}) = \arg\min_{f(\mathbf{x})} E[L(y,f(\mathbf{x}))|\mathbf{x}]
\end{equation}
For a given dataset, $\mathbf{x}\in\mathbb{R}^{d}$ is an instance or a sample. Using an additive strategy similar to other "boosting" paradigm, the functions $f(\mathbf{x})$ can be learned by the model: 
\begin{equation}
 \hat{f}(\mathbf{x}) = \hat{f}_{K}(\mathbf{x}) = \Sigma_{k=0}^M f_{k}(\mathbf{x})
\end{equation}
where $f_{0}(\mathbf{x})$ is an initial guess and $\{f_{k}(\mathbf{x})\}_{1}^{K}$ are incremental functions. $\hat{f}(\mathbf{x})=\hat{f}_{K}(\mathbf{x}):\mathbb{R}^{d}\to\mathbb{R}$ is the objective function of the model. $K$ is the number of training iterations, which also equals to the number of boosted trees. Then the function $f_{k}(\mathbf{x})$ contains the structure of the tree and leaf scores, which is a weak classification model obtained at the $k$-th training iteration. In general, the tree boosting can be defined as the objective function $\mathcal{L}$ with a training loss term and regularization term:
\begin{equation}
 \mathcal{L} = \Sigma_{i=1}^{N} l(y_{i}, \Sigma_{k=0}^M f_{k}(\mathbf{x}_{i}))+\Sigma_{k=0}^{M} \Omega(f_k) 
\end{equation}
where $N$ is the number of samples and $l$ is the logarithmic loss for multi-class classification in our scenario. To take advantages of feature vectors: $\mathbf{x}^{D}$, $\mathbf{x}^{ET}$, and $\mathbf{x}^{EN}$, we concatenate vectors from different views into a new vector $\mathbf{x}=\mathbf{x}^{D}\|\mathbf{x}^{ET}\|\mathbf{x}^{EN}$  Then the single-view classification models can be applied straightforwardly on the concatenated vector. The symbol $\|$ denotes concatenation operation on vectors from views. 

In practice, we exploit two effective versions of gradient boosting algorithms: XGBoost\footnote{https://xgboost.readthedocs.io/en/latest/} and LightGBM\footnote{https://github.com/Microsoft/LightGBM}. XGBoost has been proposed to use a second-order approximation by Taylor expansion of the loss function for the problem optimization~\cite{chen2016xgboost}. LightGBM can obtain a quite accurate estimation with smaller data size and fewer numbers of feature to speed up the conventional GBDT. Particularly, the specific gradient boosting algorithm in LightGBM we used is also GBDT. Through feature combinations across the given three views, multiple GBDT classifiers are trained independently.   

\begin{table*}[t]
\tabcolsep 0.04in
\renewcommand\arraystretch{1.3}
%\footnotesize
%\scriptsize
\caption{The details of feature combination for XGBoost models.}
\begin{tabular}{|l|c|c|c|}
\hline
\multirow{2}{*}{\textbf{Model ID}} & \multicolumn{3}{c|}{\cellcolor[gray]{0.8}\textbf{Feature Combination}}\\
\cline{2-4}
&\cellcolor[gray]{0.8} \textbf{Document View} &\cellcolor[gray]{0.8} \textbf{Entity Text View} &\cellcolor[gray]{0.8} \textbf{Entity Name View}\\
\hline
\multirow{4}{*}{\textbf{GBDT\_1}} & \multicolumn{1}{c|}{n/v/adj./adv. counts} & \multicolumn{1}{c|}{n-gram+NMF} & \multicolumn{1}{c|}{word embedding}\\
&n/v/adj./adv. counts+NMF &  &character-level encoding\\
& n-gram+NMF &  & \\
&bioentity counts &  & \\
\hline
\multirow{4}{*}{\textbf{GBDT\_2}} & \multicolumn{1}{c|}{paragraph vector} & \multicolumn{1}{c|}{sentence-level TFIDF+SVD} & \multicolumn{1}{c|}{word embedding}\\
&sentence-level TFIDF+SVD &  & character-level encoding\\
&term frequency+LDA &  & \\
&bioentity/keywords counts &  & \\
\hline
\multirow{4}{*}{\textbf{GBDT\_3}} & \multicolumn{1}{c|}{n/v/adj./adv. counts} & \multicolumn{1}{c|}{sentence-level TFIDF+SVD} & \multicolumn{1}{c|}{word embedding}\\
&n/v/adj./adv.+NMF&  & \\
&sentence-level TFIDF+SVD &  & \\
&keywords counts &  & \\
\hline
\multirow{3}{*}{\textbf{GBDT\_4}} & \multicolumn{1}{c|}{n/v/adj./adv. TFIDF} & \multicolumn{1}{c|}{n/v/adj./adv. TFIDF} & \multicolumn{1}{c|}{word embedding}\\
&sentence-level TFIDF+SVD&  &character-level encoding \\
&bioentity counts &  & \\
\hline
\end{tabular}
\label{table:featureXGBoost}
\end{table*}

\begin{table*}[t]
\tabcolsep 0.04in
\renewcommand\arraystretch{1.3}
\caption{The details of feature combination for LightGBM models.}
\begin{tabular}{|l|c|c|c|}
\hline
\multirow{2}{*}{\textbf{Model ID}} & \multicolumn{3}{c|}{\cellcolor[gray]{0.8}\textbf{Feature Combination}}\\
\cline{2-4}
&\cellcolor[gray]{0.8} \textbf{Document View} &\cellcolor[gray]{0.8} \textbf{Entity Text View} &\cellcolor[gray]{0.8} \textbf{Entity Name View}\\
\hline
\multirow{3}{*}{\textbf{GBDT\_5}} & \multicolumn{1}{c|}{n/v/adj./adv. counts} & \multicolumn{1}{c|}{n/v/adj./adv. TFIDF} & \multicolumn{1}{c|}{word embedding}\\
&n/v/adj./adv. counts+NMF &  & \\
&n/v/adj./adv. TFIDF &  & \\
\hline
\multirow{4}{*}{\textbf{GBDT\_6}} & \multicolumn{1}{c|}{n-gram+NMF} & \multicolumn{1}{c|}{n-gram+NMF} & \multicolumn{1}{c|}{word embedding}\\
&n/v/adj./adv. counts &  & character-level encoding\\
&n/v/adj./adv. counts+NMF &  & \\
&bioentity counts &  & \\
\hline
\multirow{4}{*}{\textbf{GBDT\_7}} & \multicolumn{1}{c|}{n/v/adj./adv. TFIDF} & \multicolumn{1}{c|}{n/v/adj./adv. TFIDF+SVD} & \multicolumn{1}{c|}{word embedding}\\
&n/v/adj./adv. counts+NMF&  & \\
&sentence-level TFIDF+SVD &  & \\
&n-gram+NMF &  & \\
\hline
\multirow{3}{*}{\textbf{GBDT\_8}} & \multicolumn{1}{c|}{sentence-level TFIDF+SVD} & \multicolumn{1}{c|}{sentence-level TFIDF+SVD} & \multicolumn{1}{c|}{word embedding}\\
&n/v/adj./adv. counts&  &character-level encoding  \\
&n/v/adj./adv. counts+NMF &  & \\
&keywords counts &  & \\
\hline
\multirow{3}{*}{\textbf{GBDT\_9}} & \multicolumn{1}{c|}{n/v/adj./adv. TFIDF} & \multicolumn{1}{c|}{n/v/adj./adv. TFIDF} & \multicolumn{1}{c|}{word embedding}\\
&sentence-level TFIDF+SVD&  &character-level encoding \\
&bioentity counts &  & \\
\hline
\end{tabular}
\label{table:featureLightGBM}
\end{table*}

\section{Model Ensembles}
\label{sec:4}
Many existing successful machine learning stories on challenge solutions demonstrated that combining multiple models together can gain better performance than a single model~\cite{NetflixChallenge, Gesture}. The rationale behind our framework is to combine features mined from original documents, entity texts, and entity names by different level features to form inputs of prediction models, and thus we can get numerous prediction results from these models (See Fig.~\ref{fig:framework}). By setting a threshold to the logarithmic loss score~\cite{ensemble}, $9$ qualified models finally beat other models in the comparisons. Table~\ref{table:featureXGBoost} and~\ref{table:featureLightGBM} show the feature combinations used in training these models by XGBoost and LightGBM respectively. Based on the results of basic models, ensemble strategies of $2$ models, $3$ models, and $9$ models are applied. Through model ensemble, the system can eventually output a probability distribution over classes for each sample.

Formally, let $\hat{p}^{v}_{ij}$ be the final prediction result of validation data for sample $i$ of label $j$ and $\hat{p}^{s}_{ij}$ be the final prediction result of testing data for sample $i$ on label $j$. They are computed by the linear combination of results of single models as: 

\begin{equation}
\begin{aligned}
\hat{p}^{v}_{ij} = \Sigma_c \alpha_c \hat{p}^{v}_{ijc}, \qquad \alpha_c>0\\
\hat{p}^{s}_{ij} = \Sigma_c \alpha_c \hat{p}^{s}_{ijc}, \qquad \alpha_c>0 
\end{aligned}
\end{equation}
where $\hat{p}^{v}_{ijc}$ and $\hat{p}^{s}_{ijc}$ are the predicted probability of validation data and testing data by $c$-th single model. $i$ is the index of triplet $<g_{i}, m_{i}, d_{i}>$. $j$ is the index of class. $\alpha_{c}$ is the linear combination parameter for the $c$-th model, which is a positive weight.   

Ensemble parameters $\alpha_{c}$ are computed by different manners: brute force grid searching and logarithmic loss minimization. The force grid searching quantizes the coefficient values in the interval $[0, 1]$ at increments of $0.01$. It is an efficient way to find $\alpha$ when we need to ensemble $2$ or $3$ models. On the other hand, the logarithmic loss minimization problem on validation data can be mathematically defined as:
\begin{equation}
\alpha = \arg\min_{\alpha} \text{Logloss}(\Sigma_{c}\alpha_{c}\hat{p}^{v}_{c}) 
\label{equation:minimization}
\end{equation}
Followed the evaluation metric, the $\text{Logloss}$ in our minimization problem is defined by:
\begin{equation}
\text{Logloss} = -\frac{1}{N}\Sigma_{i=1}^{N}\Sigma_{j=1}^{M} y_{ij}log(p_{i j})
\label{equation:Logloss}
\end{equation}
where $N$ and $M$ are respectively the number of triplet $<g_{i}, m_{i}, d_{i}>$ observations and the number of class labels. As we can see, Eq.~(\ref{equation:Logloss}) is consistent with the evaluation metric in Eq.~(\ref{equation:metric}) provided by the Challenge. One limitation of the ensemble method is that it treats the classes with equal importance. However, after statistical analysis, we find that the $9$ classes are severely imbalanced. In order to overcome this limitation, we compute the loss on each class to optimize its own weight $\alpha_{cj}$. Based on the Eq.~(\ref{equation:minimization}) and~(\ref{equation:Logloss}), the Logloss is updated by:
\begin{equation}
\text{Logloss} = -\Sigma_{j=1}^{M}\frac{1}{N_j} \Sigma_{i=1}^{N_j}y_{ij}log(p_{ij})
\label{eq:LLC}
\end{equation}
where $N_{j}$ is the number of triplet observations in the class $j$. The new Logloss can help us to learn weight $\alpha_{cj}$ for different classes and different models. Based on the improved ensemble method, we conduct a $9$ ensemble model. 

\section{Experimental Results}
\label{sec:5}

\subsection{Experimental Settings}
In the empirical study, we apply two offline test strategies. The first strategy is the stage-$1$ splitting which divides the entire samples into $3,321$ training samples and $368$ validation samples as shown in Table~\ref{table:datastats}; the second strategy is $5$-fold cross validation on the overall $3,689$ samples. To verify the effectiveness, the evaluation metric logarithmic loss is used, which have been introduced in Section~\ref{sec:2}.

\subsection{Effects of Multi-View Features}
In our method, features mainly come from three different views. To test the effectiveness of single feature, XGBoost implementation is utilized. In Table~\ref{table:SingleFeature}, the features are fed into the $9$ basic gradient boosting models. Their dimensions and performance on $5$-fold cross-validation are shown. We test various bag-of-word and bag-of-n-gram features with or without dimension reduction methods, and there are $15$ winner features in total built on three views. In each view, the most effective single feature can be easily observed. 

To compare two feature combinations of two views, we concatenate the features obtained by the same extraction methods, e.g., two feature vectors of term frequency+LDA are computed based on original documents and entity texts, respectively. Then we train GBDT models to test multi-view features by XGBoost implementation. Experimental results are presented in Table~\ref{table:2Combines}. Same feature derived from both document view and entity text view consistently outperforms the one only generated from a single view. The empirical study can demonstrate the effectiveness of using a complementary view.

\begin{table*}[t]
\tabcolsep 0.16in
\renewcommand\arraystretch{1.3}
\caption{The dimensions and logarithmic loss scores obtained by the single feature in $3$ views on $5$-fold cross-validation (The classifier is implemented based on XGBoost).}
\begin{tabular}{l|c|c|c}
\hline
\cellcolor[gray]{0.8} \textbf{Views} & \cellcolor[gray]{0.8} \textbf{Feature} & \cellcolor[gray]{0.8} \textbf{Dimension} & \cellcolor[gray]{0.8} \textbf{5-fold cv}  \\
\hline
\multirow{9}{*}{\textbf{Document View}} & \multicolumn{1}{c|}{bioentity counts} & \multicolumn{1}{c|}{10,022} & \multicolumn{1}{c}{0.9914}\\
&keyword counts &3,379 &\textbf{0.9583}\\
&Doc2Vec &400 &1.0037\\
&sentence-level TFIDF+SVD &100  & 0.9939 \\
&n/v/adj./adv. counts &9,868 &1.0018\\
&n/v/adj./adv. TFIDF &9,868  & 0.9825\\
&n/v/adj./adv. counts+NMF &60  & 1.0417 \\
&n-gram+NMF &60  & 1.0370 \\
&term frequency+LDA   &50  & 1.0348 \\
\hline
\multirow{4}{*}{\textbf{Entity Text View}} & \multicolumn{1}{c|}{sentence-level TFIDF+SVD} & \multicolumn{1}{c|}{200} & \multicolumn{1}{c}{0.9815}\\
&n/v/adj./adv. TFIDF  &9,868   &\textbf{0.9788} \\
&n/v/adj./adv. TFIDF+SVD &200  &1.0055 \\
&n-gram+NMF &120  &1.0029 \\
\hline
\multirow{2}{*}{\textbf{Entity Name View}} & \multicolumn{1}{c|}{word embedding} & \multicolumn{1}{c|}{200} & \multicolumn{1}{c}{\textbf{0.9811}}\\
&character-level encoding &40  & 1.1031 \\
\hline
\end{tabular}
\label{table:SingleFeature}
\end{table*}

\begin{table}[htb]
\tabcolsep 0.2in
\renewcommand\arraystretch{1.3}
\caption{Result comparisons of feature generated from single view and double views on $5$-fold cross-validation. The double view contains document view and entity text view (The classifier is implemented based on XGBoost).}
\begin{tabular}{lcc}
\hline
\cellcolor[gray]{0.8}\textbf{Feature}    \qquad\qquad     &\cellcolor[gray]{0.8}\textbf{Single View}    \qquad\qquad\qquad     &\cellcolor[gray]{0.8}\textbf{Double Views}\\
\hline
n/v/adj./adv. TFIDF  \qquad\qquad  &\textbf{0.9825} \qquad\qquad\qquad &\textbf{0.8558}\\
sentence-level TFIDF+SVD   \qquad\qquad     &0.9939   \qquad\qquad\qquad       &0.8845 \\
n/v/adj./adv. counts+NMF \qquad\qquad  &1.0417 \qquad\qquad\qquad &0.9029 \\
n/v/adj./adv. TFIDF+SVD \qquad\qquad  &1.0055$^*$ \qquad\qquad\qquad &0.8775 \\
term frequency+LDA  \qquad\qquad  & 1.0348 \qquad\qquad\qquad &0.9098 \\
\hline
\end{tabular}
$^*$ the score is based on feature in entity text view while others are computed in document view  
\label{table:2Combines}
\end{table}

\subsection{Results of Basic Models}
In the competition, $9$ different models are used in the model ensemble. Corresponding to the feature settings presented in Table~\ref{table:featureXGBoost} and~\ref{table:featureLightGBM}, Table~\ref{table:ResultsXGBOOST} and~\ref{table:ResultsLightGBM} show the results of basic gradient boosting models. For a fair comparison, all the models share the same setting of hyper-parameters. From the results, we can observe that GBDT models trained using XGBoost overall performs slightly better than those trained using LightGBM. Among the trained basic models using XGBoost, GBDT\_3 has the best performance as a single model on $5$-fold cross-validation, while GBDT\_2 has the best performance on stage-$1$ testing set. The situation for LightGBM is that GBDT\_7 is superior to other models on $5$-fold cross-validation while GBDT\_9 outperforms other models on stage-$1$ testing set.

\begin{table}[t]
\tabcolsep 0.375in
\renewcommand\arraystretch{1.3}
.\caption{Results of GBDT model in terms of logarithmic loss on $5$-fold cross-validation and stage-$1$ testing set}
\begin{tabular}{lcc}
\hline
\textbf{Model Id}    \qquad\qquad     &\textbf{5-fold cv}    \qquad\qquad\qquad     &\textbf{Stage-1 test}\\
\hline
GBDT\_1   \qquad\qquad        &0.7068      \qquad\qquad\qquad       &0.5997 \\
GBDT\_2   \qquad\qquad        &0.6930     \qquad\qquad\qquad       &\textbf{0.5638}\\
GBDT\_3   \qquad\qquad        &\textbf{0.6870}  
   \qquad\qquad\qquad       &0.5743\\
GBDT\_4   \qquad\qquad        &0.6901
   \qquad\qquad\qquad       &0.5657\\
\hline
\end{tabular}
\centering
\label{table:ResultsXGBOOST}
\end{table}

\begin{table}[t]
\tabcolsep 0.375in
\renewcommand\arraystretch{1.3}
\caption{Results of GBM model in terms of logarithmic loss on $5$-fold cross-validation and stage-$1$ testing set.}
\begin{tabular}{lcc}
\hline
\textbf{Model Id}    \qquad\qquad       &\textbf{5-fold cv}   \qquad\qquad\qquad     &\textbf{Stage-1 test}\\
\hline
GBDT\_5   \qquad\qquad        &0.7005    \qquad\qquad\qquad       &0.6090\\
GBDT\_6   \qquad\qquad        &0.7121    \qquad\qquad\qquad       &0.6152\\
GBDT\_7   \qquad\qquad        &\textbf{0.6967}    \qquad\qquad\qquad       &0.6139\\
GBDT\_8   \qquad\qquad        &0.7028    \qquad\qquad\qquad       &0.6178\\
GBDT\_9   \qquad\qquad        &0.7001    \qquad\qquad\qquad       &\textbf{0.6006}\\
\hline
\end{tabular}
\centering
\label{table:ResultsLightGBM}
\end{table}

\subsection{Results of Model Ensemble}
\begin{table}[t]
\tabcolsep 0.105in
\renewcommand\arraystretch{1.3}
\caption{Results of 2 models ensemble by brute forcing grid search.}
\begin{tabular}{llccc}
\hline
\textbf{Model\_1 Id}    \qquad\qquad  &\textbf{Model\_2 Id}    \qquad\qquad      &\textbf{weight\_1}    \qquad\qquad    &\textbf{weight\_2} \qquad\qquad       &\textbf{5-fold cv} \\
\hline
GBDT\_1   \qquad\qquad  &GBDT\_4   \qquad\qquad    &0.4   \qquad\qquad    &0.6   \qquad\qquad     &0.6786  \\
GBDT\_6   \qquad\qquad  &GBDT\_7   \qquad\qquad    &0.4    \qquad\qquad   &0.6   \qquad\qquad   &0.6846  \\
GBDT\_1   \qquad\qquad  &GBDT\_7   \qquad\qquad    &0.4    \qquad\qquad   &0.6   \qquad\qquad   &\textbf{0.6762}  \\
\hline
\end{tabular}
\label{table:2ensembleBF}
\end{table}

\begin{table}[t]
\tabcolsep 0.07in
\renewcommand\arraystretch{1.3}
\caption{Results of 3 models ensemble by brute forcing grid search.}
\begin{tabular}{lllcccc}
\hline
\textbf{Model\_1 Id}  \qquad  &\textbf{Model\_2 Id} \qquad &\textbf{Model\_3 Id}   \qquad &\textbf{weight\_1}  \qquad &\textbf{weight\_2} \qquad  &\textbf{weight\_3} \qquad  &\textbf{5-fold cv} \\
\hline
GBDT\_1  \qquad   &GBDT\_2  \qquad  &GBDT\_4 \qquad   &0.32   \qquad    &0.30  \qquad    &0.38  \qquad  &0.6738 \\
GBDT\_5   \qquad  &GBDT\_6  \qquad  &GBDT\_7  \qquad   &0.40   \qquad   &0.32  \qquad    &0.28  \qquad  &0.6818 \\
GBDT\_1   \qquad  &GBDT\_4  \qquad  &GBDT\_5  \qquad   &0.30    \qquad   &0.38  \qquad    &0.32 \qquad  &\textbf{0.6695} \\
\hline
\end{tabular}
\label{table:3ensembleBF}
\end{table}

\begin{table}[t]
\tabcolsep 0.105in
\renewcommand\arraystretch{1.3}
\caption{Results of 2 models ensemble by logarithmic loss minimization.}
\begin{tabular}{llccc}
\hline
\textbf{Model\_1 Id}    \qquad\qquad  &\textbf{Model\_2 Id}    \qquad\qquad      &\textbf{weight\_1}    \qquad\qquad     &\textbf{weight\_2} \qquad\qquad       &\textbf{5-fold cv} \\
\hline
GBDT\_1   \qquad\qquad  &GBDT\_4   \qquad\qquad   &0.49   \qquad\qquad    & 0.51   \qquad\qquad     &0.6796  \\
GBDT\_6   \qquad\qquad  &GBDT\_7   \qquad\qquad   &0.49   \qquad\qquad &0.51    \qquad\qquad   &0.6860  \\
GBDT\_1   \qquad\qquad  &GBDT\_7   \qquad\qquad   &0.49   \qquad\qquad   &0.51    \qquad\qquad   &\textbf{0.6771}  \\
\hline
\end{tabular}
\label{table:2ensembleLLM}
\end{table}

\begin{table}[t]
\tabcolsep 0.07in
\renewcommand\arraystretch{1.3}
\caption{Results of 3 models ensemble by logarithmic loss minimization.}
\begin{tabular}{lllcccc}
\hline
\textbf{Model\_1 Id} \qquad &\textbf{Model\_2 Id} \qquad &\textbf{Model\_3 Id}   \qquad &\textbf{weight\_1}  \qquad &\textbf{weight\_2} \qquad &\textbf{weight\_3} \qquad &\textbf{5-fold cv} \\
\hline
GBDT\_1   \qquad  &GBDT\_3   \qquad &GBDT\_4  \qquad  &0.33  \qquad   &0.33  \qquad   &0.33  \qquad  &0.6745 \\
GBDT\_6   \qquad  &GBDT\_8  \qquad  &GBDT\_9  \qquad  &0.33  \qquad   &0.33  \qquad   &0.33 \qquad  &0.6832 \\
GBDT\_1   \qquad  &GBDT\_4   \qquad &GBDT\_7  \qquad  &0.32  \qquad   &0.30 \qquad   &0.38   \qquad  &\textbf{0.6718} \\
\hline
\end{tabular}
\label{table:3ensembleLLM}
\end{table}

\begin{table}[t]
\tabcolsep 0.4in
\renewcommand\arraystretch{1.3}
\caption{Results of the ensemble 9 models by logarithmic loss minimization.}
\begin{tabular}{lll}
\hline
\textbf{Ensemble Method}    \qquad\qquad   &\textbf{stage-1 test}  &\textbf{5-fold cv}  \\
\hline
LogLoss\_Min       \qquad\qquad    &0.5547   &0.6711 \\
LogLoss\_Min\_cl   \qquad\qquad    &\textbf{0.5506} &\textbf{0.6694}    \\
\hline
\end{tabular}
\label{table:improvedEns}
\end{table}

\begin{figure}
\centering
\includegraphics[width=1\textwidth]{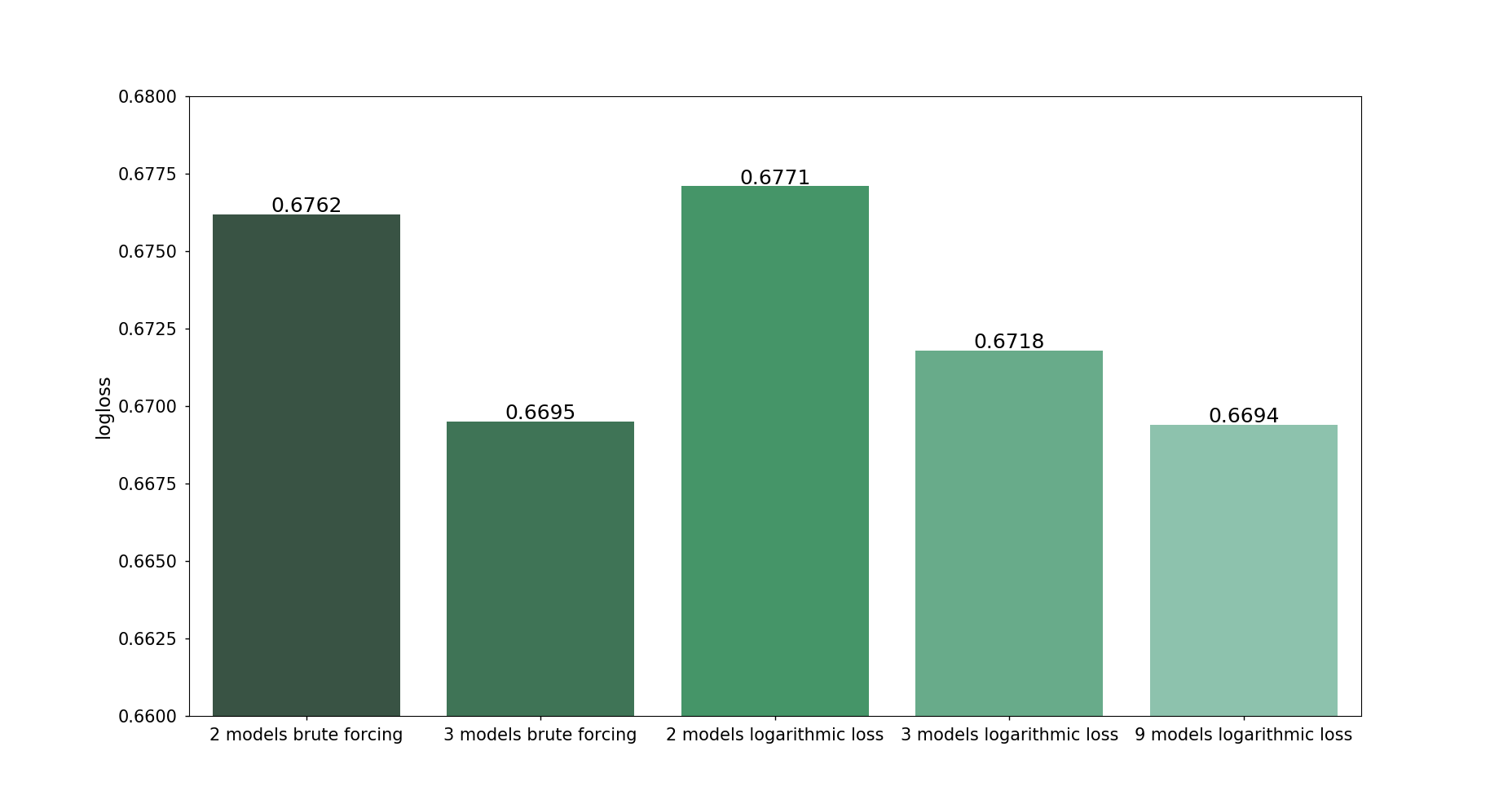}
\caption{\label{fig:EnsembleResult} Experimental results of different model ensemble strategies on 5-fold cross validation.}
\end{figure}

Similarly, 5-fold cross-validation to the model ensemble is utilized here. In practice, brute force gird search strategy and logarithmic loss minimization strategy are used in the model ensemble. The combinations of basic models are shown in tables, if the evaluation of Logloss scores are less than a threshold. Table~\ref{table:2ensembleBF} and~\ref{table:3ensembleBF} respectively show ensemble results as well as weights by brute force grid search strategy to ensemble 2 models and 3 models. 

Table~\ref{table:2ensembleLLM} and~\ref{table:3ensembleLLM} respectively show 2 and 3 ensemble results under the target of logarithmic loss minimization. The best model ensemble can be found in the results. The improved logarithmic loss minimization considering the imbalanced labels are also tested by $5$-fold cross-validation. The results in Table~\ref{table:improvedEns} show that the improved ensemble strategy can increase prediction accuracy on ensemble results of $9$ models. To compare the ensemble effects of $9$ models to $2$ models and $3$ models, the Fig.~\ref{fig:EnsembleResult} plots the Logloss scores of main model ensemble methods concerned in this paper. Among different strategies, $9$ model ensemble is the final winner, which slightly outperforms the $3$ model ensemble based on brute forcing grid search.        

\section{Conclusion}
The main contribution of our work is developing a comprehensive pipeline to perform gene mutation classification based on clinical articles. Our solution mines text features from three views including original document view, entity text view, and entity name view. Various machine learning algorithms are exploited to generate text features from perspectives of domain knowledge, document-level, sentence-level, and word-level. In addition, word embedding and character-level encoding based on entity names are adopted. Multiple GBDT classifiers with different feature combinations are utilized in ensemble learning to achieve a satisfying classification accuracy. The reported results demonstrate that our multi-view ensemble classification framework yields promising performances in this competition. 

\section*{Acknowledgement}
The work is partially supported by NSF IIS-1650723, IIS-1716432 and IIS-1750326. The authors would like to thank the support from Amazon Web Service Machine Learning for Research Award (AWS MLRA).

\bibliographystyle{plain}
\bibliography{refs}

\begin{thebibliography}{10}

\bibitem{AC}
Alfred~V Aho and Margaret~J Corasick.
\newblock Efficient string matching: an aid to bibliographic search.
\newblock {\em Communications of the ACM}, 18(6):333--340, 1975.

\bibitem{NetflixChallenge}
Robert~M. Bell and Yehuda Koren.
\newblock Lessons from the netflix prize challenge.
\newblock {\em SIGKDD Explor. Newsl.}, 9(2):75--79, 2007.

\bibitem{LDA}
David~M Blei, Andrew~Y Ng, and Michael~I Jordan.
\newblock Latent dirichlet allocation.
\newblock {\em Journal of machine Learning research}, 3(Jan):993--1022, 2003.

\bibitem{chen2016xgboost}
Tianqi Chen and Carlos Guestrin.
\newblock Xgboost: A scalable tree boosting system.
\newblock In {\em Proceedings of the 22nd acm sigkdd international conference
  on knowledge discovery and data mining}, pages 785--794. ACM, 2016.

\bibitem{wordnet}
Christiane Fellbaum.
\newblock {\em WordNet}.
\newblock Wiley Online Library, 1998.

\bibitem{friedman2001greedy}
Jerome~H Friedman.
\newblock Greedy function approximation: a gradient boosting machine.
\newblock {\em Annals of statistics}, pages 1189--1232, 2001.

\bibitem{SVD}
Gene~H Golub and Charles~F Van~Loan.
\newblock {\em Matrix computations}, volume~3.
\newblock JHU Press, 2012.

\bibitem{harris}
Zellig~S Harris.
\newblock Distributional structure.
\newblock {\em Word}, 10(2-3):146--162, 1954.

\bibitem{he2005microrna}
Lin He, J~Michael Thomson, Michael~T Hemann, Eva Hernando-Monge, David Mu,
  Summer Goodson, Scott Powers, Carlos Cordon-Cardo, Scott~W Lowe, Gregory~J
  Hannon, et~al.
\newblock A microrna polycistron as a potential human oncogene.
\newblock {\em nature}, 435(7043):828--833, 2005.

\bibitem{huang2011genetukit}
Minlie Huang, Jingchen Liu, and Xiaoyan Zhu.
\newblock Genetukit: a software for document-level gene normalization.
\newblock {\em Bioinformatics}, 27(7):1032--1033, 2011.

\bibitem{postagging}
Dan Jurafsky and James~H Martin.
\newblock {\em Speech and language processing}, volume~3.
\newblock Pearson London:, 2014.

\bibitem{doc2vec}
Quoc Le and Tomas Mikolov.
\newblock Distributed representations of sentences and documents.
\newblock In {\em Proceedings of the 31st International Conference on Machine
  Learning (ICML-14)}, pages 1188--1196, 2014.

\bibitem{NMF}
Daniel~D Lee and H~Sebastian Seung.
\newblock Learning the parts of objects by non-negative matrix factorization.
\newblock {\em Nature}, 401(6755):788--791, 1999.

\bibitem{tfidf}
David~AC Manning and JGM Decleer.
\newblock Introduction to industrial minerals.
\newblock 1995.

\bibitem{ensemble}
Gr{\'e}goire Mesnil, Tomas Mikolov, Marc'Aurelio Ranzato, and Yoshua Bengio.
\newblock Ensemble of generative and discriminative techniques for sentiment
  analysis of movie reviews.
\newblock {\em arXiv preprint arXiv:1412.5335}, 2014.

\bibitem{Word2Vec}
Tomas Mikolov, Kai Chen, Greg Corrado, and Jeffrey Dean.
\newblock Efficient estimation of word representations in vector space.
\newblock {\em arXiv preprint arXiv:1301.3781}, 2013.

\bibitem{skip-gram}
Tomas Mikolov, Ilya Sutskever, Kai Chen, Greg~S Corrado, and Jeff Dean.
\newblock Distributed representations of words and phrases and their
  compositionality.
\newblock In {\em Advances in neural information processing systems}, pages
  3111--3119, 2013.

\bibitem{Tutorials}
Nanyun Peng, Hoifung Poon, Chris Quirk, Kristina Toutanova, and Wen-tau Yih.
\newblock Cross-sentence n-ary relation extraction with graph lstms.
\newblock {\em arXiv preprint arXiv:1708.03743}, 2017.

\bibitem{GloVe}
Jeffrey Pennington, Richard Socher, and Christopher Manning.
\newblock Glove: Global vectors for word representation.
\newblock In {\em Proceedings of the 2014 conference on empirical methods in
  natural language processing (EMNLP)}, pages 1532--1543, 2014.

\bibitem{DimReduction}
Sam~T Roweis and Lawrence~K Saul.
\newblock Nonlinear dimensionality reduction by locally linear embedding.
\newblock {\em science}, 290(5500):2323--2326, 2000.

\bibitem{wei2013tmvar}
Chih-Hsuan Wei, Bethany~R Harris, Hung-Yu Kao, and Zhiyong Lu.
\newblock tmvar: a text mining approach for extracting sequence variants in
  biomedical literature.
\newblock {\em Bioinformatics}, 29(11):1433--1439, 2013.

\bibitem{gennorm}
Chih-Hsuan Wei and Hung-Yu Kao.
\newblock Cross-species gene normalization by species inference.
\newblock {\em BMC bioinformatics}, 12(8):S5, 2011.

\bibitem{wei2013pubtator}
Chih-Hsuan Wei, Hung-Yu Kao, and Zhiyong Lu.
\newblock Pubtator: a web-based text mining tool for assisting biocuration.
\newblock {\em Nucleic acids research}, 41(W1):W518--W522, 2013.

\bibitem{Gesture}
Jiaxiang Wu, Jian Cheng, Chaoyang Zhao, and Hanqing Lu.
\newblock Fusing multi-modal features for gesture recognition.
\newblock In {\em Proceedings of the 15th ACM on International Conference on
  Multimodal Interaction}, pages 453--460, 2013.

\bibitem{textclassification}
Xiang Zhang, Junbo Zhao, and Yann LeCun.
\newblock Character-level convolutional networks for text classification.
\newblock In {\em Advances in neural information processing systems}, pages
  649--657, 2015.

\end{thebibliography}

\end{document}